\documentclass[journal]{IEEEtran}
\usepackage{multirow} 
\usepackage{amsmath,amsfonts}
\usepackage{algorithmic}
\usepackage{algorithm}
\usepackage{array}
\usepackage{booktabs}
\usepackage[caption=false,font=normalsize,labelfont=sf,textfont=sf]{subfig}
\usepackage{textcomp}
\usepackage{stfloats}
\usepackage{url}
\usepackage{verbatim}
\usepackage{graphicx}
\usepackage{cite}
\usepackage{amsmath}
\usepackage[table,xcdraw]{xcolor}
\usepackage{amssymb}
\usepackage{xcolor}
\definecolor{rblue}{rgb}{0,0.5,1}
\definecolor{hollywoodcerise}{rgb}{0.96, 0.0, 0.63}
\definecolor{lasallegreen}{rgb}{0.03, 0.47, 0.19}
\definecolor{hanpurple}{rgb}{0.32, 0.09, 0.98}
\definecolor{green(pigment)}{rgb}{0.0, 0.65, 0.31}

\usepackage[pagebackref=false,breaklinks=true,colorlinks,bookmarks=false]{hyperref}
\hypersetup{colorlinks,linkcolor={red},citecolor={hanpurple},urlcolor={magenta}}  
\DeclareUnicodeCharacter{2061}{}

\begin{document}

\title{
HierDAMap: Towards Universal Domain Adaptive BEV Mapping via Hierarchical Perspective Priors
}

\vspace{-2em}
\author{Siyu Li$^{1,2}$, Yihong Cao$^{3}$, Hao Shi$^{4}$, Yongsheng Zang$^{5}$, Xuan He$^{6}$, Kailun Yang$^{1}$, and Zhiyong Li$^{1,5}$%
\thanks{This work was supported in part by the National Natural Science Foundation of China (Grant No. U23A20341, No. 61976086, and No. 62473139), in part by the Hunan Provincial Research and Development Project (Grant No. 2025QK3019, No.2025QK1005), and in part by the State Key Laboratory of Autonomous Intelligent Unmanned Systems (the opening project number ZZKF2025-2-10).
\textit{(Corresponding authors: Zhiyong Li and Kailun Yang.)}}
\thanks{$^{1}$The authors are with the School of Artificial Intelligence and Robotics and the National Engineering Research Center of Robot Visual Perception and Control Technology, Hunan University, Changsha, China (email: zhiyong.li@hnu.edu.cn; kailun.yang@hnu.edu.cn).}
\thanks{$^{2}$The author is also with the School of Automation and Electrical Engineering, Zhejiang University of Science and Technology, Hangzhou, China.}
\thanks{$^{3}$The author is with the Key Laboratory of Big Data Research and Application for Basic Education, Hunan Normal University, Changsha, China.}
\thanks{$^{4}$The author is with Ant Group, Beijing, China.}
\thanks{$^{5}$The authors are with the College of Computer Science and Electronic Engineering, Hunan University, Changsha, China.}
\thanks{$^{6}$The author is with Hunan Vanguard Group Corporation Limited, Changsha, China.}
}

\maketitle

\begin{abstract}
The exploration of Bird's-Eye View (BEV) mapping technology has driven significant innovation in visual perception technology for autonomous driving. BEV mapping models need to be applied to the unlabeled real world, making the study of unsupervised domain adaptation models an essential path. However, research on unsupervised domain adaptation for BEV mapping remains limited and cannot perfectly accommodate all BEV mapping tasks. To address this gap, this paper proposes HierDAMap, a universal and holistic BEV domain adaptation framework with hierarchical perspective priors. Unlike existing research that solely focuses on image-level learning using prior knowledge, this paper explores the guiding role of perspective prior knowledge across three distinct levels: global, sparse, and instance levels. With these priors, HierDAMap consists of three essential components, including Semantic-Guided Pseudo Supervision (SGPS), Dynamic-Aware Coherence Learning (DACL), and Cross-Domain Frustum Mixing (CDFM). SGPS constrains the cross-domain consistency of perspective feature distribution through pseudo labels generated by vision foundation models in 2D space. To mitigate feature distribution discrepancies caused by spatial variations, DACL employs uncertainty-aware predicted depth as an intermediary to derive dynamic BEV labels from perspective pseudo-labels, thereby constraining the coarse BEV features derived from corresponding perspective features. CDFM, on the other hand, leverages perspective masks of the view frustum to mix multi-view perspective images from both domains, which guides cross-domain view transformation and encoding learning through mixed BEV labels. Furthermore, this paper introduces intra-domain feature exchange data augmentation to enhance the efficiency of domain adaptation learning. The proposed method is verified on multiple BEV mapping tasks, such as BEV semantic segmentation, high-definition semantic, and vectorized mapping. It demonstrates competitive performance across various conditions, including weather scenarios, regions, and datasets. The source code will be made publicly available at \url{https://github.com/lynn-yu/HierDAMap}.
\end{abstract}

\vspace{-0.5em}
\vspace{-1em}

\section{Introduction}
\begin{figure}[tb]
      \centering
      \includegraphics[scale=0.5]{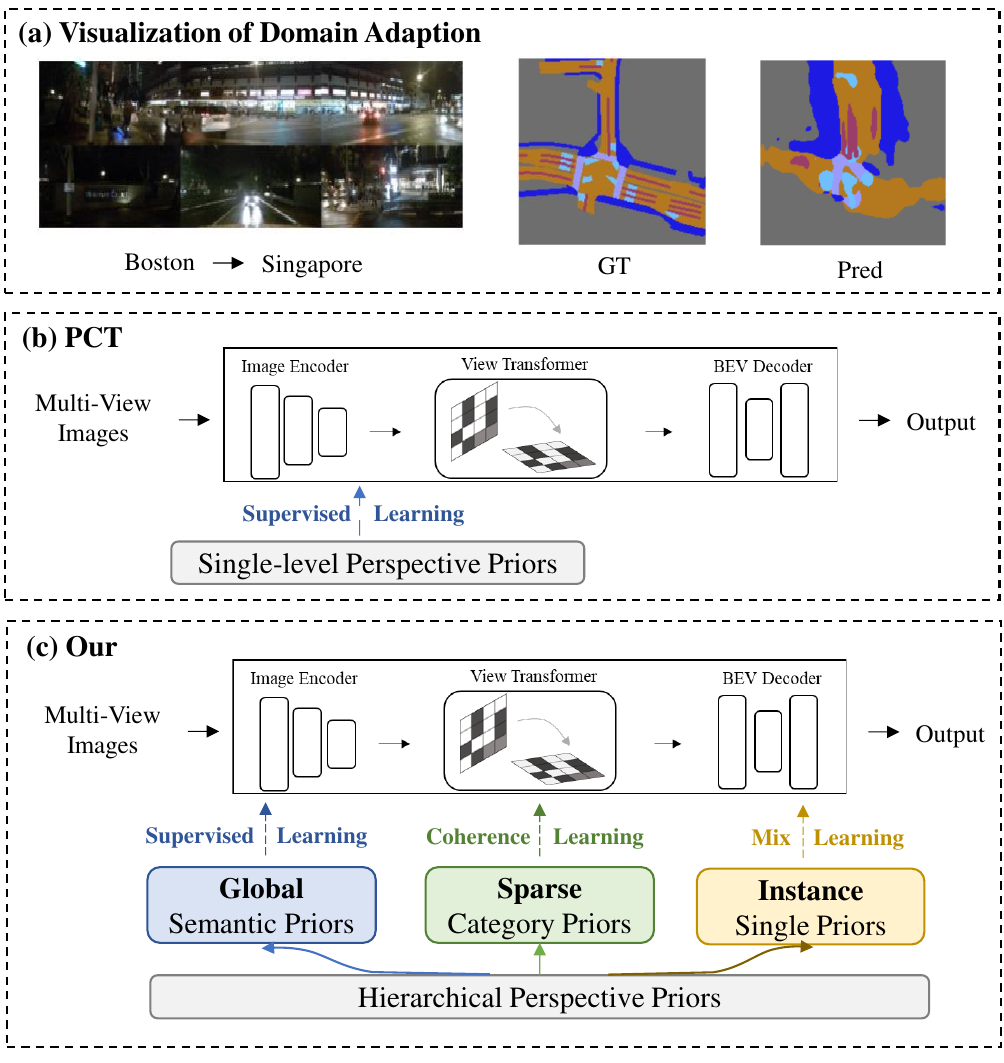}
      \caption{Visual illustration and technical framework of BEV models in different domains. 
      (a) presents the visualization of the model prediction in Singapore while the model is trained on the Boston dataset.
      The predicted results of the model trained on Boston are unsatisfactory due to large domain gaps.
      (b) shows the framework of the representative previous domain adaptation paradigm PCT~\cite{pct}. 
      It only employs perspective pseudo-label supervision in the whole domain at the image coding level. 
      (c) depicts our framework. The perspective priors are hierarchically fully exploited to promote domain adaptation at different levels of the BEV mapping model.
      }
      \label{Fig.1}
\end{figure}
Bird's-Eye-View (BEV), a plane view perpendicular to the visual perspective,  has accelerated the development of end-to-end models for perception and planning in autonomous driving~\cite{bevsurvey}.
Recently, research in BEV understanding has leaped forward in different tasks, such as semantic segmentation~\cite{LSS,ibevseg}, object detection~\cite{BEVFormer,bevdepth}, and map construction~\cite{hdmapnet,maptr}. 

Undeniably, the majority of research focuses on fully supervised datasets, resulting in poor model performance when encountering unseen environments.
As shown in Fig.~\ref{Fig.1}-(a), when the training and testing regions differ, the BEV mapping model fails to accurately depict environmental information. 
Given the potential domain gaps between data from different regions, BEV mapping models need to explore more robust domain transfer capabilities, which is also a necessary research direction for unsupervised real-world applications.
There is limited research on the domain adaptation of BEV mapping.
Moreover, due to the flourishing domain adaptation in perspective view tasks, most efforts focus on enhancing the adaptation capabilities of perspective modules in the current limited research. However, BEV mapping models also require learning transformations across different views, making it challenging to apply these methods across all BEV mapping tasks.

\begin{figure}[tb]
      \centering
      \includegraphics[scale=0.35]{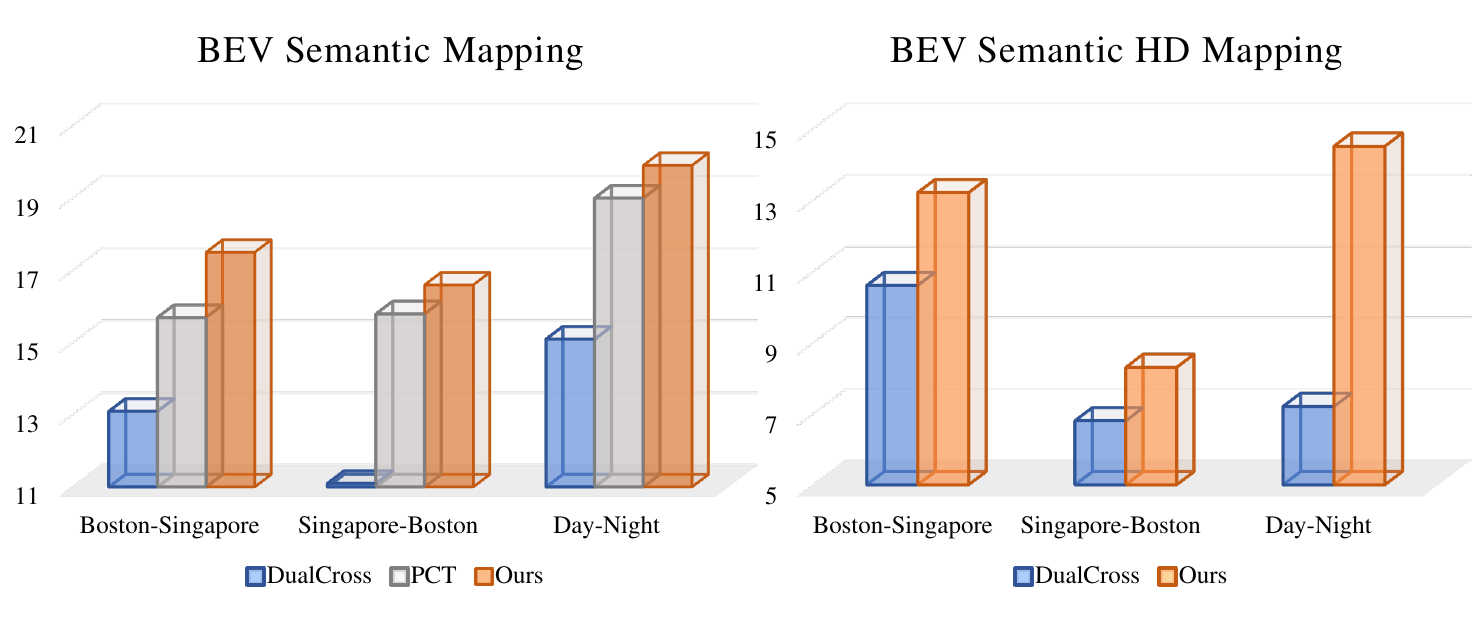}
      \caption{Results of different Unsupervised Domain Adaptation (UDA) methods for BEV mapping. Our method shows superior performance across various cross-domain scenarios for BEV mapping. The state-of-the-art methods DualCross~\cite{man2023dualcross} and PCT~\cite{pct} are compared.
      }
      \label{fig.inta}
\end{figure}
DualCross~\cite{man2023dualcross}, as the first domain adaptation study for BEV mapping, employed adversarial learning~\cite{adversarial} to achieve domain adaptation at the perspective and decoding levels, respectively.
PCT~\cite{pct}, on the other hand, explored the guidance of the perspective priors to domain adaptation. 
They leverage perspective prior knowledge to implement supervision at the image encoding level, as shown in Fig.~\ref{Fig.1}-(b).
However, applying global supervision solely at the image encoding layer is insufficient for BEV tasks. 
The BEV mapping framework roughly consists of multiple steps: the image encoder, the view transformer, and the BEV decoder.
Although global supervision of image encoding can provide reliable image features across different domains, the cross-domain learning of instance spatial relationships is limited in effectiveness.
Therefore, building on the coupled structure of BEV models, this paper designs a hierarchical perspective prior-guided domain adaptation framework that better accommodates various BEV mapping tasks, such as semantic mapping and semantic High-Definition (HD) mapping, as shown in Fig.~\ref{fig.inta}.

Concretely, we propose HierDAMap, a holistic BEV domain adaptation learning framework with hierarchical perspective priors, to handle various BEV mapping tasks in a unified way.
Additionally, perspective prior knowledge, derived from generalized vision foundation models, can provide pseudo-label knowledge for the unseen domain.
The whole framework consists of three modules: Semantic-Guided Pseudo Supervision (SGPS), Dynamic-Aware Coherence Learning (DACL), and Cross-Domain Frustum Mixing (CDFM).
They are distributed across three levels--global, sparse, and instance--to progressively achieve BEV domain adaptation through perspective priors.
Firstly, SGPS is proposed to ensure strong generalization of image features through supervising the encoding module.  
Then, conditioned on perspective pseudo-labels and estimated depth, sparse BEV pseudo-labels are obtained through the view transformer in DACL. 
These pseudo-labels with dynamic awareness are used to supervise the coarse BEV features generated through the transformation. 
Finally, CDFM  utilizes the instance mask groups belonging to each perspective image to mix source and target domain images while generating mixed labels based on the BEV frustum range corresponding to the perspective views, which can guide the transformation learning in the target domain. 
Furthermore, a feature exchange data augmentation module is designed to improve the efficiency of domain adaptation learning.
We evaluate the method under multiple settings and different tasks assembled by nuScenes~\cite{nus} and Argoverse~\cite{argoverse} datasets. 
Extensive experiments show that our model has state-of-the-art performance in various cross-domain BEV mapping tasks.

The main contributions delivered in this work are summarized as follows:
\begin{itemize}
    \item We propose HierDAMap, a universal domain adaptation framework with hierarchical perspective priors for various BEV map construction tasks. 
    \item Based on perspective prior knowledge, pseudo semantic information effectively supervises image encoding, while dynamic labels in BEV space constrain semantic consistency during the view transformer, and the mixing of frustum instances across domains guides BEV feature generation.
    \item Our method outperforms previous domain adaptation models facing different BEV tasks in various experimental settings, including cross-scene and cross-dataset domain shift scenarios.
\end{itemize}

\section{Related Work}
\label{related}
\noindent \textbf{BEV Mapping:}
BEV mapping tasks focus on modeling static objects of environments, such as lanes, zebra crossings, and stop lines. Benefiting from the high cost-effectiveness of cameras, camera-based BEV mapping has become a prominent area of research in recent studies.
The core of this research lies in how to extract 3D spatial features from a 2D perspective image.
LSS~\cite{LSS} estimated the depth distribution of perspective images to project the 2D features into the 3D space coupled with the camera parameters. 
BEVDepth~\cite{bevdepth} leveraged LiDAR depth to supervise depth estimation, which can improve the reliability of depth information and construct an accurate BEV map. CoBEV~\cite{shi2024cobev} combined depth and height cues to construct robust BEV features. BEVPool~\cite{bevpoolv2} designed a lightweight view transformation method for faster inference. 

While the previous methods generate corresponding features by projecting from 2D to 3D, the subsequent approaches capture the relevant features from 3D to 2D.
BEVFormer~\cite{BEVFormer, BEVFormerv2} was a typical work in this area. It initialized the spatial representation using a grid of uniform 3D points. Similarly, it utilized camera parameters to project these points onto the 2D perspective view, capturing the corresponding features. In practice, this projection relationship is fixed. Therefore, GKT~\cite{gkt} designed an indexing table, significantly improving the projection speed.
In addition to research on view transformation, BEV mapping also involves the design of task-specific detection heads.
HDMapNet~\cite{hdmapnet} first proposed a framework for online High-Definition (HD) mapping, including semantic mapping, instance detection, and direction detection tasks. BEVSegFormer~\cite{bevsegformer} was also part of the research on HD semantic mapping, where a Transformer-based architecture was used for decoding and learning.
VectorMapNet~\cite{vecmapnet}, MapTR~\cite{maptr,maptrv2}, StreamMapNet~\cite{streammapnet}, and InstaGraM~\cite{InstaGraM} explored lightweight HD mapping and vectorized mapping, where each instance consisted of vector points and lines. 
However, these models, trained with full supervision on datasets, have suboptimal mapping results when they are directly applied in real-world environments.
Since real-world environments lack labels, unsupervised domain adaptation learning is a necessary approach to enhance model performance.

\noindent\textbf{Perspective Domain Adaptation:}
Research on domain adaptation has become increasingly mature in perspective-based tasks, with many representative works. 
Some approaches~\cite{adv-1,adversarial,ganlearn} focused on leveraging Generative Adversarial Network (GAN)~\cite{adv} to align cross-domain features.
The work~\cite{ganlearn} proposed a self-supervised adversarial network for pavement distress classification, where a pretext module was designed to mine the foreground region for feature alignment.
Another part of the work opted for a self-training approach based on pseudo-labels from Mean Teacher (MT)~\cite{MT}.
Specifically, pseudo-labels generated from the original data were used to supervise the results of data-augmented learning in the student model.
The work~\cite{transmt} introduced a domain encoding module to exploit the specific features of each domain.

Efficient data augmentation methods, such as CutMix~\cite{cutmix}, dropout~\cite{dropout,unimatch}, and camera dropout~\cite{pct}, can effectively guide domain adaptation learning.
To alleviate the problem that pseudo-labels are difficult to learn fine structures, MIC~\cite{mic} proposed the mask consistency learning module,  which leverages spatial contextual relationships as additional cues to guide domain adaptation learning.
MICDroup~\cite{micdrop} combined depth features and employed a bidirectional masking approach to learn the contours of visual features, thereby generating more accurate pseudo-labels.
The work of~\cite{maskvideo} applied similarity theory to the study of video domain adaptation and explored a high-quality fusion of self-training and feature adversarial learning.
In addition, some works~\cite{dacs,openmix,mix-2} improve the quality of pseudo-labels by using mixed learning.
Guided by the instance segmentation results of pseudo-labels, DACS~\cite{dacs} fused the instance image patches corresponding to the source domain to the target image. 
In contrast to the former, where patches are of fixed position and size, the work~\cite{openmix} designed a random mixture of position and size, which helps to accurately predict the shape of unknown classes.
Furthermore, the work~\cite{mix-2} implemented cross-domain blending in pixels and used a contrastive learning method to constrain feature learning.
The aforementioned domain mixing methods are all unidirectional, lacking the reverse information that transfers target-domain objects into the source domain. To address this limitation, ADPL~\cite{ADPL} proposed a dual-path domain mixing scheme, which learns in parallel along two complementary paths to improve the quality of target-domain pseudo-labels.
QuadMix~\cite{UDASS} further designed four-directional mixing paths, which not only bridge the cross-domain feature distributions but also reduce intra-domain feature distances, thereby achieving more comprehensive domain alignment.

With the great success of universal vision-language models, \textit{e.g.,} CLIP~\cite{clip}, in image classification tasks, they have also been widely applied in pixel-level semantic segmentation tasks.
Based on instance masks obtained from the Segment Anything model~\cite{sam}, SAN~\cite{san} combines the CLIP model to identify the semantic categories of each mask. 
It is an open-vocabulary semantic segmentation model, which will also be applied in this work to provide semantic pseudo-labels for perspective views.
Domain adaptation for BEV tasks differs from that of perspective image tasks. 
Unlike perspective image tasks, which focus on feature learning within a single view, BEV involves the transformation between two distinct views, making cross-domain learning particularly challenging.

\begin{figure*}[tb]
      \centering
      \includegraphics[scale=0.5]{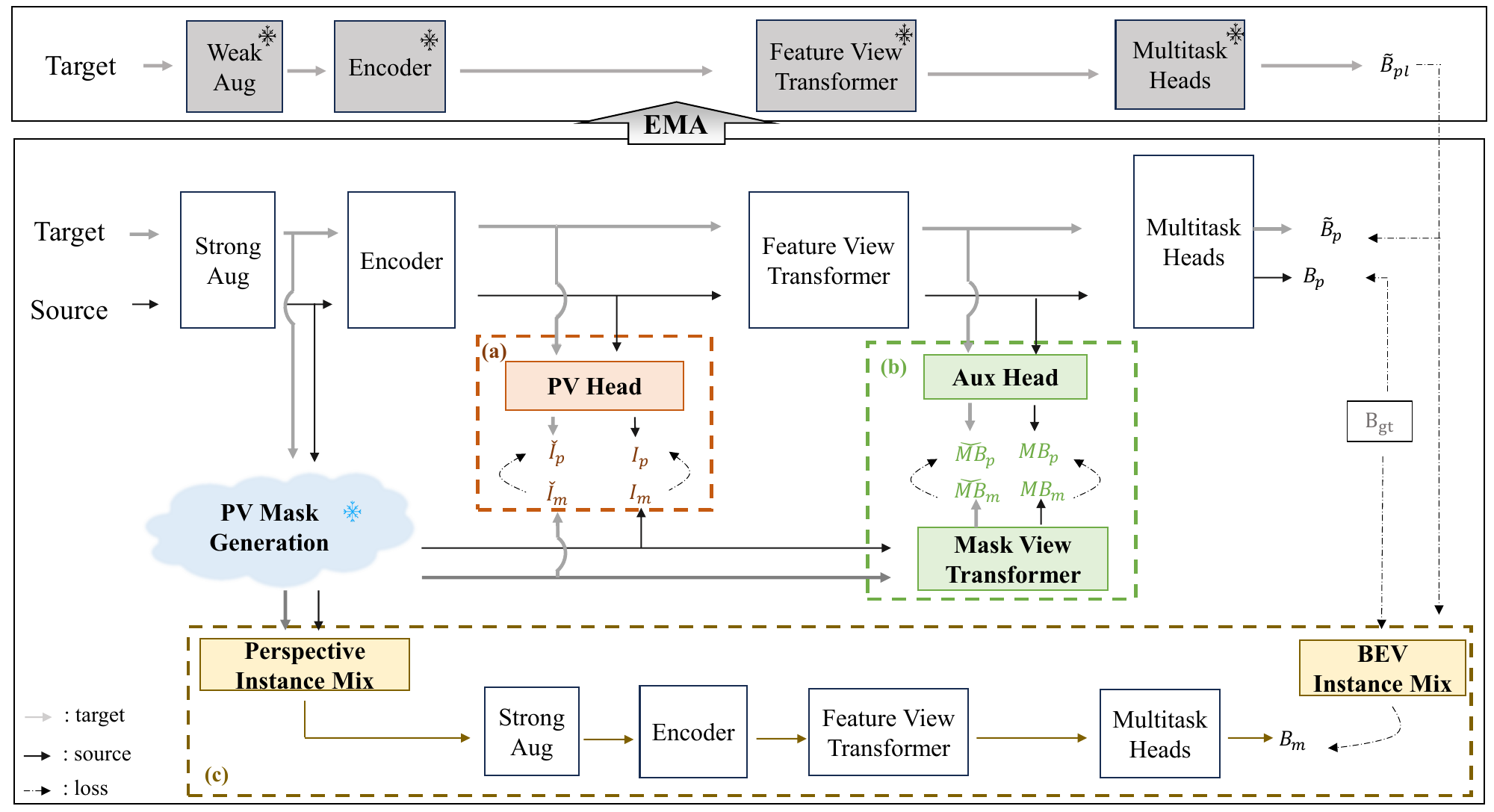}
      \caption{The framework of HierDAMap. The entire framework is based on mean teacher, where the student model parameters are learned from both the source and target domains, whereas the teacher model dynamically adjusts based on the student model changes, which is controlled by parameter $\alpha$.
      }
      \label{Fig.domainframe}
\end{figure*}

\noindent\textbf{Visual BEV Domain Adaptation:}
Previous research on BEV models has primarily focused on improving accuracy, while the study of adaptation capabilities should not be overlooked.
Some studies~\cite{semvecnet,genmapping} directly focused on improving the generalization performance of the model. 
Semi-supervised learning is also relevant to domain adaptation, where there is much research~\cite{skyeye,s2g2,semibev,explore} on BEV tasks.
Most studies focused on monocular BEV, with work~\cite{semibev} proposing a data augmentation method that synchronously distorts the perspective view and BEV. 
However, this augmentation approach was unsuitable for multi-view BEV tasks due to perspective differences, which could lead to misalignment between the BEV space and perspective space features. This paper proposes a data augmentation method at the BEV global feature level for multi-view image tasks, which can effectively enhance domain adaptation capabilities.

Recently, some research works have explored BEV scene understanding from the domain adaptation perspective.
Similar to domain adaptation of perspective views, domain adaptation for BEV tasks can be roughly categorized into two types: one leverages adversarial learning to guide feature consistency~\cite{man2023dualcross,dabev}, whereas the other employs self-training with pseudo-labels~\cite{pct,bevuda,semibev}.
DualCross~\cite{man2023dualcross} proposed a multi-modal cross-domain adaptation framework. It not only designed adversarial learning at the image and BEV feature level but also proposed point cloud distillation to improve feature generation robustness.
DABEV~\cite{dabev} presented query-based designs and exploited image-view features or BEV features to regularize the adaptation of the other.
BEVUDA~\cite{bevuda} designed a three-level consistency learning based on pseudo-label guidance, which the domain discriminator realizes.
PCT~\cite{pct} explored a BEV domain adaptation framework, where perspective pseudo labels are essential cues to supervise perspective features. 
However, they ignore the equal importance of geometric spatial relationships in BEV models. It is worth exploring how to make full use of perspective pseudo-labels to improve BEV domain learning. 
This paper utilizes hierarchical perspective prior knowledge to construct a unified domain adaptation BEV mapping model.

\section{HierDAMap: Proposed Framework}
In this work, we focus on unsupervised domain adaptation for BEV mapping. 
We propose HierDAMap, a holistic unsupervised domain adaptation framework based on hierarchical perspective priors to address different domain-adaptive BEV mapping tasks in a unified way.
First, we introduce the overall framework of HierDAMap in Sec.~\ref{sec:framework_hierdamap}. 
Then, we provide a brief description of the BEV mapping model in Sec.~\ref{sec:bev_mapping_model}.
Finally, we elaborate on the domain adaptation module guided by hierarchical prior knowledge in Sec.~\ref{sec:pipeline_hierarchical_perspective_priors}. Simultaneously, a data augmentation method tailored for BEV tasks is introduced in Sec.~\ref{sec:feature_exchange_data_augmentation}.

\subsection{Framework of HierDAMap}
\label{sec:framework_hierdamap}
The domain adaptation framework proposed in this paper comprises a teacher-student model, which is based on a mean teacher architecture, as illustrated in Fig.~\ref{Fig.domainframe}.  
The structures of the teacher model and the student model are identical, consisting of a BEV mapping model that includes an image encoder, a view transformer, and multitask decoder modules.
Based on the parameters of the student model, the parameters of the teacher model are updated through the Exponential Moving Average (EMA) mode.

The student model completes learning combined with the source domain BEV labels $B_{gt}$, the target domain pseudo-labels $\widetilde{B}_{pl}$, and the perspective pseudo-labels of the full domain $I_m$.
Among these, the target domain pseudo-labels are generated by the teacher model based on weakly augmented data.
The perspective pseudo-labels are produced from the large-scale vision foundation model.
With the development of vision-based models, the task of perspective semantic segmentation has developed rapidly~\cite{light,hyper}.  
Meanwhile, the accuracy of domain adaptation in perspective views continues to improve. 
Particularly with the advent of large vision foundation models, the generalization capability of perspective view tasks has reached new heights. 
Based on the segment anything model~\cite{sam}, the Side Adapter Network (SAN)~\cite{san} integrated the analysis and comprehension capabilities of the CLIP language model~\cite{clip} to further generate semantic tags for the masks.
Considering the strong domain adaptability of this model, HierDAMap leverages SAN to generate perspective masks in different scenes.
Note that this perspective mask is generated offline and is utilized exclusively during the training phase.

By hierarchically guiding through perspective prior knowledge at three levels—global, sparse, and instance—three domain adaptation learning modules are additionally designed to achieve joint domain adaptation learning of semantic and geometric features in the BEV model. These will be detailed in the subsequent sections, as illustrated by modules (a), (b), and (c) in Fig.~\ref{Fig.domainframe}. 

\subsection{BEV Mapping Model}
\label{sec:bev_mapping_model}
\begin{figure*}[tb]
      \centering
      \includegraphics[scale=0.6]{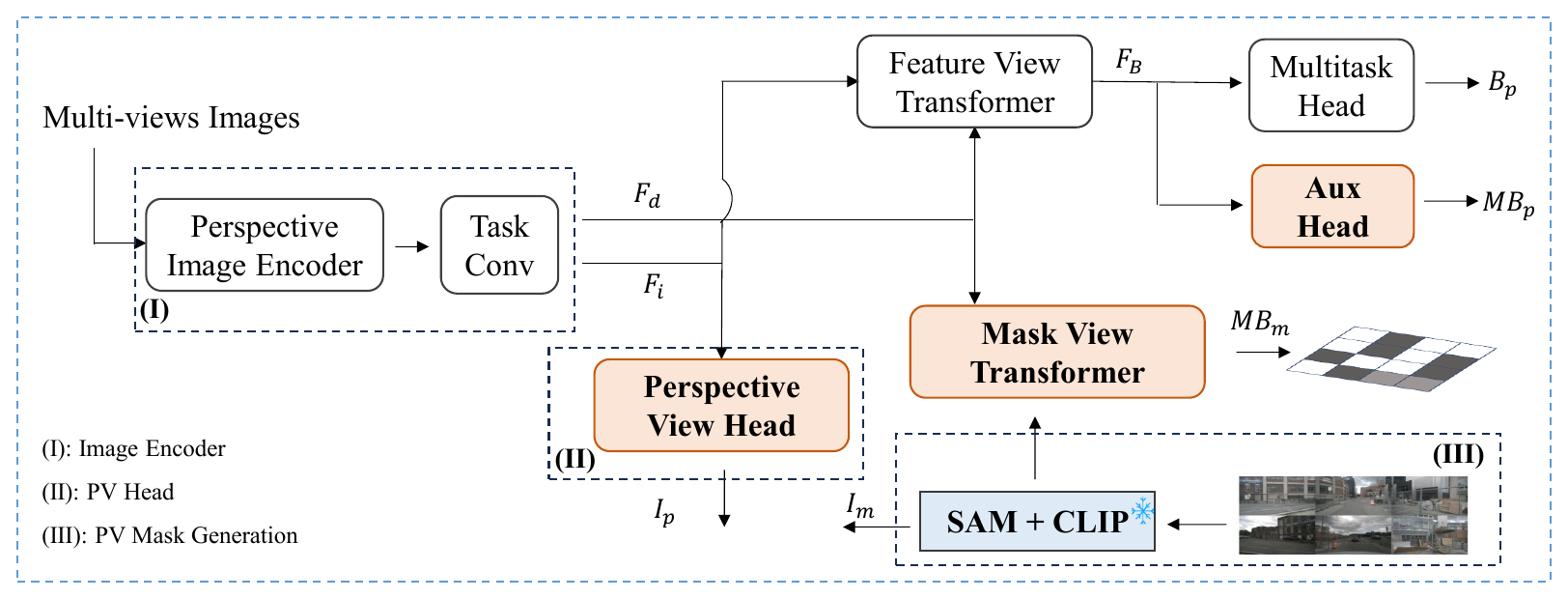}
      \caption{Illustration of the BEV mapping model. Based on the LSS framework~\cite{LSS}, it contains five modules: Image Encoder, Perspective View Head (PV Head), Perspective View Mask Generation (PV Mask Generation), View Transformer, and Multitask Head.
      }
      \label{Fig.frame}
\end{figure*}

The core of the BEV mapping model lies in the transformation between two perspectives. 
The Lift, Splat, Shoot (LSS) approach~\cite{LSS} achieves this transformation by integrating depth estimation with camera parameters. 
This method not only performs well in fully supervised mapping tasks~\cite{maptrv2} but is also widely applied in domain adaptation and generalization research~\cite{pct,semvecnet}.
Therefore, this work leverages the robust LSS as the foundational mapping model to design a universal domain adaptation framework, as shown in Fig.~\ref{Fig.frame}.
Specifically, multi-view perspective images $I_i$ ($i \in [0,n]$, $n$ is the number of images) as input data are first passed through the image encoder to generate deep perspective features $F_{i}$ and depth estimates $F_{d}$. 
\begin{equation}
    F_i, F_d = Encoder(I_j).
\end{equation}
$D_{p}$ is the probability that each pixel belongs to a certain depth range within the set range:
\begin{equation}
    D_p = SoftMax(F_d),
\end{equation}
Combined with intrinsic parameters $P_{in}$, extrinsic parameters $P_{ex}$, and image pre-processing transformation parameters $P_t$, the coarse BEV features $F_B$ can be obtained. For subsequent computational efficiency, these parameters are expanded to a size of $h\times w \times 3 \times 3$. 
\begin{equation}
\label{vt}
    F_B = VT(F_i,D_p,P_{in},P_{ex},P_t),
\end{equation}
where $VT$ is the feature view transformer module. 
Finally, the BEV mapping $B_p$ can be obtained from the BEV decoder.
Additionally, to align with the multi-level domain adaptation learning pipeline, this paper presents three newly designed components: a perspective view head, mask view transformer, and an auxiliary head, which are detailed as follows.

\subsection{Proposed Architecture of Hierarchical Perspective Priors}
\label{sec:pipeline_hierarchical_perspective_priors}

From the BEV mapping model, it is evident that BEV features are derived from both the semantic and geometric features of perspective views. 
Semantic understanding models for perspective views exhibit strong domain adaptation capabilities. Therefore, we propose a domain adaptation pipeline that guides BEV domain adaptation learning through hierarchical perspective prior knowledge, thereby transferring the robust domain adaptation capabilities of perspective knowledge to BEV mapping tasks.
Here, this paper designs three guided learning modules, \textit{i.e.} Semantic-Guided Pseudo Supervision (SGPS), Dynamic-Aware Coherence Learning (DACL), and Cross-Domain Frustum Mixing (CDFM).

\subsubsection{Semantic-Guided Pseudo Supervision }
Intuitively, perspective priors provide robust semantic information. This information can indirectly constrain the generation of BEV features by directly supervising perspective view features across different domains. 
Therefore, this section assembles this semantic information into reliable pseudo-labels $I_m$ to supervise the full-domain learning of perspective feature encoding.
The semantic information is derived from a large-scale vision foundation model, SAN~\cite{san}. 

Specifically, an efficient perspective view head $H$ is designed to learn perspective semantic segmentation $I_p$. 
This head is consistent with the non-bottleneck module referenced from ERFNet~\cite{erfnet}, which maintains high precision while improving learning efficiency.
\begin{equation}
    I_p = H(F_i).
\end{equation}

Dice loss as the supervised loss $Dice$ is leveraged in this task. 
It is important to note that pseudo-labels $I_m$ are available for both the source domain and the target domain.
\begin{equation}
    Loss_p = Dice(I_p,I_m).
\end{equation}

Certainly, through the supervised learning of a global perspective view, the model can achieve consistent semantic feature distribution across domains, thereby providing a robust foundation for subsequent BEV features.

\subsubsection{Dynamic-Aware Coherence Learning}
The geometric relationships of 2D and 3D are also a crucial component of BEV models. 
In general, BEV models utilize geometric relationships to convert perspective features into BEV features $F_b$, as demonstrated in Eq.~\ref{vt}. 
Although perspective features constrain prototype semantics through supervision with pseudo labels, the unconstrained depth estimation during view transformer may disrupt the semantic representation of the prototypes, thereby generating inaccurate BEV features. 
This uncertainty poses a significant obstacle to learning in the target domain where labeled supervision is absent. 
Consequently, we explore whether reducing such uncertainty can enhance cross-domain generalization performance. 
We have designed a dynamic label mechanism aimed at reducing uncertainty by maintaining the consistency of prototype features before and after the view transformer.

Dynamic labels $MB_m$ are generated through a mask view transformer, which uses reliable perspective pseudo-labels and learnable depth estimates. 
As shown in Fig.~\ref{Fig.frame}, the pseudo-labels come from the perspective view mask generation module, while depth estimates are from the image encoder module.
Firstly, since the depth estimates for each view are high-resolution, the perspective pseudo-labels are first resized to match the dimensions of the depth estimates. 
Subsequently, the mask view transformer module is employed to project these high-resolution pseudo-labels into the BEV space, following the same computational process as outlined in Eq.~\ref{vt}. 
The key distinction lies in the replacement of perspective features $F_i$ with resized pseudo-labels $I_{m}^{'}$. Additionally, the depth estimates are represented by a one-hot encoding $\theta_o$ rather than being computed probabilistically, which is to generate unique dynamic labels $MB_m$.
\begin{equation}
    MB_m = VT(I_{m}^{'},\theta_o(D_p),P_{in},P_{ex},P_t).
\end{equation}

Given the unique semantics of the dynamic labels, an auxiliary task head, consisting of activation function layers, normalization layers, and convolution layers, is introduced to learn the supervised feature $MB_p$ for these labels. 
The constraint is implemented through a loss $Loss_y$.
\begin{equation}
    Loss_y = ||MB_p-MB_m||_2.
\end{equation}

\subsubsection{Cross-Domain Frustum Mixing}
In the domain adaptation learning tasks for perspective views~\cite{classmix,dacs}, it has been demonstrated that mixed learning from both the source and target domains can enhance the generalization capabilities. 
However, domain mixing methods for perspective views are difficult to directly apply to BEV tasks, primarily because it is challenging to achieve a one-to-one correspondence between perspective instance masks and BEV instance labels. This implies that BEV tasks require a tailored domain mixing solution.
In this section, unlike the approach of domain mixing at the level of individual instances, we design a domain mixing scheme based on instance groups of a single perspective, taking advantage of the unique view frustum that each perspective view has in BEV space.
Within this context, source domain instances located above the road surface within a single view frustum can guide the learning of road space in the target domain.

Given that all mapping instances belong to the ground plane, vehicles, as three-dimensional entities above the ground, not only interact with these instances but also play a crucial role in enhancing the understanding of the geometric structure within the environmental context.
Therefore, we have selected vehicles as the domain mixing objects. 
Geometric spatial relationships in the target domain can be guided with the help of the supervised detection of vehicle instances in the source domain. 
Based on instance masks $Inst_m$ derived from perspective pseudo-labels, we integrate vehicle masks of each perspective image from the source domain into the target domain, generating composite multi-view images $I_{j}^{'}$. In the integration process $M_x$, target domain pixels identified as belonging to the source domain instance masks are replaced by their corresponding source domain instance pixels.
\begin{equation}
    I_{j}^{'} = M_x(Inst_m,I_j^{s},I_j^{t}),
\end{equation}

Subsequently, BEV maps and vehicle detections are generated within the mixed domain by processing these composite images through the BEV mapping model. 
Due to the inconsistent image pre-processing matrices for the source $T^{s}$ and target domains $T^{t}$, the matrices within the view transformer module also need to be mixed. 
\begin{equation}
    T^{'} = M_x(Inst_m,T^{s},T^{t}),
\end{equation}
\begin{equation}
    F_b^{'} = VT(F_i^{'},D_p^{'},P_{in},P_{ex},P_{t}^{'}),
\end{equation}
where $M_x$ represents a mixed function measured in pixel units.

After obtaining the mixed pred $B_m$, the learning of this module is supervised by labels that are a blend of the source domain ground truth $B_{gt}$ and the target domain pseudo-labels $\tilde{B}_{pl}$, a process constrained by the loss $Loss_{mix}$.
\begin{equation}
    Loss_{mix} = \mathrm{L2}({B}_m,M_x(\tilde{B}_{pl},B_{gt})).
\end{equation}

\begin{figure}[tb]
      \centering
      \includegraphics[scale=0.65]{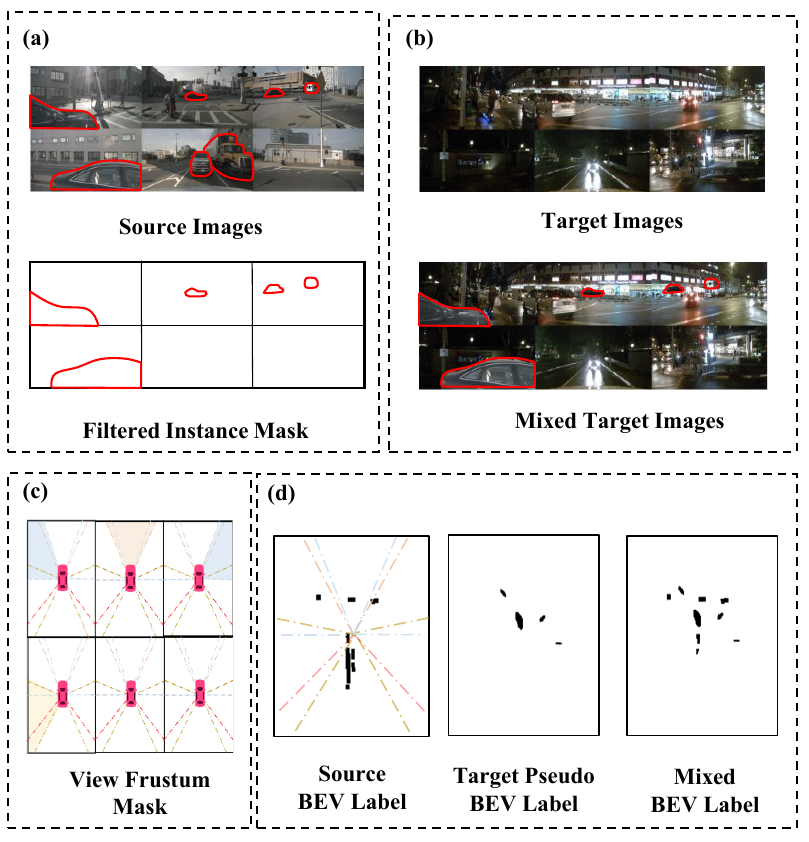}
      \caption{The example diagram of cross-domain instance mixing. (a) depicts the generation and filtration of the source instance mask. (b) depicts the generation process of mixed target perspective images, which is implemented by mixing instance masks from the source. (c) is the corresponding view frustum mask from the source filtered instance mask. The first row corresponds to the front-left, the front, and the front-right. The second row represents the back perspective views. (d) corresponds to the mixed BEV labels obtained from the view frustum mask, source and target labels.
      }
      \label{Fig.mix}
      \vspace{-2.0em}
\end{figure}

Since instance mixing does not modify map labels but only affects vehicle labels in BEV space, it is necessary to mix BEV vehicle labels from different domains.
Acknowledging the inherent inaccuracies present in perspective pseudo-labels, this paper employs an adaptive methodology to amalgamate vehicle labels across diverse perspectives. 
As shown in Fig.~\ref{Fig.mix}, if instance masks are present within a perspective view of the source domain, the BEV vehicle labels within the corresponding view frustum range are incorporated into mixed labels.
Due to inherent errors in perspective models, BEV instances and perspective instances within the same view frustum may not be perfectly aligned. Nevertheless, the supervisory signal from detected perspective instances can provide invaluable guidance for model learning.
Simultaneously, to prevent instances from excessively obscuring genuine environmental information, the view containing the highest number of occupied pixels is isolated. Consequently, the BEV instances within its corresponding view frustum are excluded from the mixing process. 
As shown in Fig.~\ref{Fig.mix}, instances from the back view that contain the highest pixel ratio of vehicles among all perspectives are excluded from domain mixing selection. Certainly, the BEV instance labels within the corresponding view frustum are also preserved from mixing.

\subsection{Feature Exchange Data Augmentation}
\label{sec:feature_exchange_data_augmentation}
Data augmentation serves as an efficient strategy for generalization learning in unlabeled target domains.
In BEV tasks, data augmentation is predominantly applied to perspective images, with limited research exploring data augmentation within the BEV space.
Consequently, this work investigates a data augmentation method based on BEV features to provide a stronger data-augmented synchronous learning pipeline for the target domain.

Given the diversity of map instances, some instances span the entire plane, while others, such as zebra crossings, are confined to localized areas. Therefore, our data augmentation strategy is designed to address both global and local considerations.
In the global strategy, two modes are designed. 
The first mode involves discarding a portion of global features through Dropout~\cite{dropout} during the decoding learning process, applied to BEV features generated from weakly augmented multi-view images.
The second entails randomly exchanging features between BEV features generated from weakly and strongly augmented data along the channel dimension, preserving feature integrity while enhancing diversity.
From a local strategy perspective, similar to the previous one, it differs in that it randomly exchanges features along the positional dimension, augmenting the spatial representation of features.
Note that the selection probability for each of the three modes is identical.
This data augmentation strategy is applied to a new training pipeline of the target domain by the loss $Loss_{da}$, as illustrated in Fig.~\ref{Fig.domainframe}.
\begin{equation}
    Loss_{da} = \mathrm{L2}(\tilde{B}_{da},\tilde{B}_{pl}),
\end{equation}

\subsection{Overall Loss}
The loss used for source domain $loss^{s}$ is as followed:
\begin{equation}
\label{loss_1}
    Loss^{s} = Loss_{gt} + \lambda_1 Loss_p^{s} + \lambda_2 Loss_s^{s} + Loss_d,
\end{equation}
\begin{equation}
    Loss_{gt} = L_{a}(B_p,B_{gt}),
\end{equation}
where $L_{a}$ is the task loss by ground truth $B_{gt}$. $Loss_d$ is the depth loss.
The loss of target domain is $Loss^{t}$:
\begin{equation}
\label{loss_2}
    Loss^{t} = \beta (Loss_{pl}+Loss_{mix}+2*Loss_{da}) +\lambda_1 Loss_p^{t} + \lambda_2Loss_s^{t},
\end{equation}
where $Loss_{pl}$ is the loss supervised by the target domain pseudo-label $B_{pl}$. $Loss_{mix}$ is the loss of cross-domain pipeline, and $Loss_{da}$ is the data augmentation loss. 
These three losses are implemented through the L2 loss between the target domain pseudo-labels and the predictions.
$\lambda_1$ and $\lambda_2$ are the loss weights. 
Finally, the overall loss is $Loss^{s} + Loss^{t}$.

\section{Experiment}
\begin{table*}[t]
\fontsize{9}{13.8}\selectfont
\renewcommand{\arraystretch}{1.1}
\setlength\tabcolsep{9pt}
\caption{Semantic HD mapping performance (IoU\%) on different UDA benchmarks.}
\vspace{-1.5em}
\label{tab:hdmapping}
\begin{center}
\resizebox{0.9\linewidth}{!}{\begin{tabular}{ccccc|ccccc}
\toprule [2pt]
\multicolumn{1}{l|}{\multirow{2}{*}{Method}} & \multicolumn{3}{c|}{IoU} & \multirow{2}{*}{mIoU}  & \multicolumn{1}{l|}{\multirow{2}{*}{Method}} & \multicolumn{3}{c|}{IoU} & \multirow{2}{*}{mIoU} \\ \cline{2-4} \cline{7-9}
\multicolumn{1}{l|}{} & Boundary & Pedestrian & \multicolumn{1}{l|}{Divider} &  & \multicolumn{1}{l|}{} & Boundary & Pedestrian & \multicolumn{1}{l|}{Divider} &  \\ \hline
\multicolumn{5}{c|}{\textbf{Boston $\xrightarrow{}$ Singapore}} & \multicolumn{5}{c}{\textbf{Dry $\xrightarrow{}$ Rain}} \\ \hline
\multicolumn{1}{l|}{Source Only} & 14.4 & 1.4 & 16.5 & \multicolumn{1}{|l|}{10.8} & \multicolumn{1}{l|}{Source Only} & 20.7 & 8.5 & 21.3 & \multicolumn{1}{|l}{16.8} \\
\multicolumn{1}{l|}{DualCross~\cite{man2023dualcross}} & 16.1 & 1.5 & 14.2 & \multicolumn{1}{|l|}{10.6} & \multicolumn{1}{l|}{DualCross~\cite{man2023dualcross}} & 20.6 & 7.6 & 22.5 & \multicolumn{1}{|l}{16.9} \\
\multicolumn{1}{l|}{Our} & 20.1 & 1.5 & 18.1 & \multicolumn{1}{|l|}{\textbf{13.2}} & \multicolumn{1}{l|}{Our} & 24.7 & 10.8 & 27.2 & \multicolumn{1}{|l}{\textbf{20.9}} \\ \bottomrule [2pt]
\multicolumn{5}{c|}{\textbf{Singapore $\xrightarrow{}$ Boston}} & \multicolumn{5}{c}{\textbf{Day $\xrightarrow{}$ Night}} \\ \hline
\multicolumn{1}{l|}{Source Only} & 12.3 & 0.00 & 7.1 & \multicolumn{1}{|l|}{6.4} & \multicolumn{1}{l|}{Source Only} & 10.4 & 0.00 & 9.1 & \multicolumn{1}{|l}{6.5} \\
\multicolumn{1}{l|}{DualCross~\cite{man2023dualcross}} & 12.4 & 0.02 & 7.7 & \multicolumn{1}{|l|}{6.8} & \multicolumn{1}{l|}{DualCross~\cite{man2023dualcross}} & 11.0 & 0.00 & 10.5 & \multicolumn{1}{|l}{7.2} \\
\multicolumn{1}{l|}{Our} & 13.5 & 0.01 & 11.2 & \multicolumn{1}{|l|}{\textbf{8.3}} & \multicolumn{1}{l|}{Our} & 20.6 & 0.00 & 22.0 & \multicolumn{1}{|l}{\textbf{14.5}}  \\ \bottomrule [2pt]
\end{tabular}}
\end{center}
\vspace {-2.0em}
\end{table*}
\begin{table*}[t]
\fontsize{4}{5}\selectfont
\renewcommand{\arraystretch}{1.1}
\setlength\tabcolsep{7pt}
\caption{Semantic mapping performance (IoU\%) on different benchmarks of the nuScenes dataset~\cite{nus}. $*$ represents data from the work~\cite{pct}}
\vspace{-2.0em}
\label{tab:smapping}
\begin{center}
\resizebox{0.9\linewidth}{!}{
\begin{tabular}{cccccccccc}
\toprule [1pt]
\multicolumn{1}{l|}{\multirow{2}{*}{Method}}  & \multicolumn{1}{l|}{\multirow{2}{*}{Image Size}} & \multicolumn{6}{c|}{IoU} & \multicolumn{1}{l}{\multirow{2}{*}{mIoU}} \\ \cline{3-8}
\multicolumn{1}{l|}{}   & \multicolumn{1}{l|}{} & Dri. & Ped. & Walk. & Stop. & Car. & \multicolumn{1}{l|}{Div.} & \multicolumn{1}{l}{}   \\ \hline
\multicolumn{9}{c}{\textbf{Boston $\xrightarrow{}$ Singapore}} \\ \hline
\multicolumn{1}{l|}{Source Only}  & \multicolumn{1}{l|}{128${\times}$352} & 40.2 & 6.3 & 7.0 & 1.9 & 0.7 & \multicolumn{1}{l|}{9.7} & \multicolumn{1}{l}{10.9}  \\

\multicolumn{1}{l|}{DualCross~\cite{man2023dualcross}}  & \multicolumn{1}{l|}{128${\times}$352} & 43.8  &8.2   &13.5   &4.5   &3.6    & \multicolumn{1}{l|}{9.6} & \multicolumn{1}{l}{13.9}  \\
\multicolumn{1}{l|}{DomainADV*~\cite{man2023dualcross}} & \multicolumn{1}{l|}{224${\times}$480} & 40.0 & 8.3 & 11.7 & 4.5 & 2.2 & \multicolumn{1}{l|}{11.6} & \multicolumn{1}{l}{13.1}  \\
\multicolumn{1}{l|}{PCT*~\cite{pct}}  & \multicolumn{1}{l|}{224${\times}$480} & 46.2 & 8.6 & 14.2 & 6.4 & 3.7 & \multicolumn{1}{l|}{15.0} & \multicolumn{1}{l}{15.7}  \\
\multicolumn{1}{l|}{Our}  & \multicolumn{1}{l|}{128${\times}$352} & 51.6 & 9.8 & 15.6 & 7.3 & 6.2 & \multicolumn{1}{l|}{14.3} & \multicolumn{1}{l}{\textbf{17.5}}  \\ \bottomrule [1pt]
\multicolumn{9}{c}{\textbf{Singapore $\xrightarrow{}$ Boston}} \\ \hline
\multicolumn{1}{l|}{Source Only}  & \multicolumn{1}{l|}{128${\times}$352} & 44.7 & 2.6 & 11.5 & 4.2 & 0.2 & \multicolumn{1}{l|}{8.4} & \multicolumn{1}{l}{11.9}  \\
\multicolumn{1}{l|}{DualCross~\cite{man2023dualcross}}  & \multicolumn{1}{l|}{128${\times}$352} & 45.7 &2.2 &13.6 &6.4 &5.2  & \multicolumn{1}{l|}{9.9} & \multicolumn{1}{l}{13.8}  \\
\multicolumn{1}{l|}{DomainADV*~\cite{man2023dualcross}}  & \multicolumn{1}{l|}{224${\times}$480} & 35.7 & 4.2 & 11.3 & 4.8 & 0.6 & \multicolumn{1}{l|}{9.7} & \multicolumn{1}{l}{11.1}  \\
\multicolumn{1}{l|}{PCT*~\cite{pct}}  & \multicolumn{1}{l|}{224${\times}$480} & 47.0 & 8.0 & 19.3 & 6.3 & 0.7 & \multicolumn{1}{l|}{13.7} & \multicolumn{1}{l}{15.8} \\
\multicolumn{1}{l|}{Our}  & \multicolumn{1}{l|}{128${\times}$352} & 50.2 & 4.9 & 19.0 & 7.5 & 3.7 & \multicolumn{1}{l|}{13.9} & \multicolumn{1}{l}{\textbf{16.6}} \\
 \bottomrule [1pt]
 \multicolumn{9}{c}{\textbf{Dry$\xrightarrow{}$ Rain}} \\ \hline
\multicolumn{1}{l|}{Source Only}  & \multicolumn{1}{l|}{128${\times}$352} & 67.1 & 29.5 & 35.8 & 23.4 & 24.6 & \multicolumn{1}{l|}{25.1} & \multicolumn{1}{l}{34.2}  \\
\multicolumn{1}{l|}{DualCross~\cite{man2023dualcross}}  & \multicolumn{1}{l|}{128${\times}$352} & 67.9 &29.8 &38.4 &22.6 &29.3 & \multicolumn{1}{l|}{25.4  } & \multicolumn{1}{l}{35.6}  \\
\multicolumn{1}{l|}{DomainADV*~\cite{man2023dualcross}} & \multicolumn{1}{l|}{224${\times}$480} & 72.0 & 39.8 & 42.0 & 33.7 & 38.9 & \multicolumn{1}{l|}{33.6} & \multicolumn{1}{l}{43.3} \\
\multicolumn{1}{l|}{PCT*~\cite{pct}} & \multicolumn{1}{l|}{224${\times}$480} & 78.3 & 45.2 & 52.1 & 37.6 & 47.2 & \multicolumn{1}{l|}{36.4} & \multicolumn{1}{l}{\textbf{49.5}} \\
\multicolumn{1}{l|}{Our}  & \multicolumn{1}{l|}{128${\times}$352} & 68.9 & 31.4 & 37.9 & 25.1 & 30.0 & \multicolumn{1}{l|}{27.4} & \multicolumn{1}{l}{36.8}  \\ \bottomrule [1pt]
\multicolumn{9}{c}{\textbf{Day $\xrightarrow{}$ Night}} \\ \hline
\multicolumn{1}{l|}{Source Only}  & \multicolumn{1}{l|}{128${\times}$352} & 32.8 & 2.2 & 4.3 & 4.4 & 0.0 & \multicolumn{1}{l|}{9.2} & \multicolumn{1}{l}{8.8}  \\
\multicolumn{1}{l|}{DualCross~\cite{man2023dualcross}}  & \multicolumn{1}{l|}{128${\times}$352} & 49.4 &8.0 &12.7 &7.9 &0.0   & \multicolumn{1}{l|}{15.1} & \multicolumn{1}{l}{15.5}  \\
\multicolumn{1}{l|}{DomainADV*~\cite{man2023dualcross}}  & \multicolumn{1}{l|}{224${\times}$480} & 37.1 & 16.4 & 10.7 & 5.7 & 0.0 & \multicolumn{1}{l|}{11.2} & \multicolumn{1}{l}{15.1}  \\
\multicolumn{1}{l|}{PCT*~\cite{pct}} & \multicolumn{1}{l|}{224${\times}$480} & 51.3 & 19.4 & 16.1 & 7.6 & 0.0 & \multicolumn{1}{l|}{19.3} & \multicolumn{1}{l}{19.0}  \\
\multicolumn{1}{l|}{Our}  & \multicolumn{1}{l|}{128${\times}$352} & 58.7 & 14.7 & 17.3 & 8.3 & 0.0 & \multicolumn{1}{l|}{20.6} & \multicolumn{1}{l}{\textbf{19.9}}  \\  \bottomrule [1pt]
\end{tabular}}
\end{center}
\vspace {-2.0em}
\end{table*}

In this section, we conduct extensive experiments to evaluate the effectiveness of the proposed HierDAMap for universal domain adaptive BEV mapping. In Sec.~\ref{Sec:experiment_settings}, we introduce the experimental setup, including the specific data configuration for unsupervised domain adaptation. Then, we present the domain adaptation research setup for different tasks in Sec.~\ref{sec:implementation_details}, where this paper investigates three BEV mapping tasks. Finally, we present the comparative results and ablation analysis.

\subsection{Experiment Setting}
\label{Sec:experiment_settings}
We verify the effectiveness of the proposed model under various cross-domain settings across two datasets, nuScenes~\cite{nus} and Argoverse~\cite{argoverse}.
On the nuScenes dataset, the cross-domain adaption performance is validated across four scenarios, following the domain gap division established in works~\cite{man2023dualcross,pct}: $Boston \xrightarrow{} Singapore$, $Singapore \xrightarrow{}Boston$, $Day \xrightarrow{} Night$, and $Dry \xrightarrow{}Rain$.

Additionally, the sensor configurations between the nuScenes and Argoverse datasets are inconsistent, presenting an additional challenge for domain adaptation learning in BEV tasks. 
The former achieves full-scene perception using six cameras, while the latter utilizes seven cameras. Additionally, the camera parameters and installation positions differ significantly.
Thus, cross-dataset adaptation learning is further validated in this study.
Overall, we will conduct experiments on three tasks under different cross-domain settings across two datasets.
All experiments are conducted on NVIDIA RTX A6000 GPUs. 

\subsection{Implementation Details}
\label{sec:implementation_details}
\textbf{Semantic HD Mapping:}
The baseline of semantic HD mapping is chosen as LSS~\cite{LSS}. 
It employs Efficient-B0~\cite{efficientnet} as the image encoder and adopts a ResNet-18 architecture~\cite{resnet} as the decoder. 
This task specifically focuses on segmenting line categories, which refers to HDMapNet~\cite{hdmapnet}, including Boundary, Pedestrian, and Divider. 
Grid maps have a resolution of $0.15m$, a size of $(400,200)$ on the nuScenes dataset, and $(200,400)$ on the Argoverse dataset. 
The training batch size is $12$, and the learning rate is set to $3e^{-3}$. The initial training epochs are $24$ for nuScenes and $6$ for Argoverse.
We use mean Intersection over Union (mIoU) as the main evaluation metric.

\textbf{Semantic Mapping:}
The model framework for this task is similar to the previous task. The static categories follow PCT~\cite{pct}, including Drivable Area, Pedestrian Crossing, Walkway, Stop Line, Carpark Area, and Divider.
The learning rate is set to $3e^{-3}$ when the training batch size is $12$.
The range of semantic mapping is $(-50m,50m)$, and the resolution is $0.5m$.
The mapping performance is measured by mean Intersection over Union (mIoU).

\textbf{Vectorized HD Mapping:}
Though the detection targets are the same as semantic HD mapping, the vectorized mapping task, describing the map objects using points and lines, is fundamentally different from the previous mapping approach. The Average Precision (AP) is adopted as the evaluation metric in this task, which is based on the Chamfer Distance (CD). 
Under three CD thresholds $\{0.5m,1.0m,1.5m\}$, the average is the final evaluation metric (mAP).
In this work, we use MapTRv2~\cite{maptrv2} as the baseline to investigate cross-domain performance. 
The size and resolution of BEV features remain the same as in the previous task.
The training batch is $4$, and the initial learning rate is $3.75e^{-4}$. 

The domain adaptation framework for all tasks is based on the mean teacher benchmark. 
The learning momentum of the teacher model is $\alpha = 0.99$. Additionally, as depicted in Fig.~\ref{Fig.domainframe}, the model is trained using both simple and strong data augmentation techniques. 
Specifically, the simple augmentation involves resize, while the strong augmentation incorporates additional transformations, including rotation, cropping, flipping, and colorjitter.
For loss weights of Eq.~\ref{loss_1} and Eq.~\ref{loss_2}, $\lambda_1=0.5$, $\lambda_2=0.01$. 
Furthermore, $\beta$ is controlled by a sigmoid ramp-up function, which starts at $0$ and gradually increases to $0.1$ when the training round is halfway through.

\subsection{Main Results}
\textbf{Semantic HD Mapping}
For semantic HD mapping, we conduct a comparison across the four UDA scenarios divided within the nuScenes dataset. Adversarial learning is a key technique in domain adaptation tasks involving scene-level adaptation. Although the work~\cite{pct} also utilized perspective priors, it has not been open-sourced. Therefore, this paper chooses the cross-modal and cross-domain adversarial learning method proposed in the DualCross~\cite{man2023dualcross} as the baseline for comparison. 
As shown in Table~\ref{tab:hdmapping}, although DualCross outperforms the baseline in most cross-domain experiments, it shows suboptimal performance in the cross-domain scenario from $Boston \xrightarrow{} Singapore$, which has mixed domain gaps. 
In contrast, our method consistently delivers superior performance across all four cross-domain scenarios, effectively showcasing its efficacy in the semantic HD mapping task.
Even in the challenging domain shift from $Day \xrightarrow{} Night$, our method demonstrates a significant improvement, outperforming the baseline by $8.0\%$.

\begin{table}[bt]
\fontsize{5}{5.8}\selectfont
\renewcommand{\arraystretch}{1.1}
\setlength\tabcolsep{9pt}
\caption{Detection accuracy of car in each UDA scenario.}
\vspace{-2.0em}
\label{tab:car_iou}
\begin{center}
\resizebox{1.0\linewidth}{!}{
\begin{tabular}{lc|lc}
\toprule [1pt]
\multicolumn{1}{l|}{Method} & IoU & \multicolumn{1}{l|}{Method} & IoU \\ \hline
\multicolumn{2}{c|}{\textbf{Boston$\xrightarrow{}$ Singapore}} & \multicolumn{2}{c}{\textbf{Singapore$\xrightarrow{}$ Boston}} \\ \hline
\multicolumn{1}{l|}{DualCross~\cite{man2023dualcross}} & 20.5 & \multicolumn{1}{l|}{DualCross~\cite{man2023dualcross}} & -- \\
\multicolumn{1}{l|}{PCT~\cite{pct}} & 19.7 & \multicolumn{1}{l|}{PCT~\cite{pct}} & -- \\
\multicolumn{1}{l|}{Our} & 23.4 & \multicolumn{1}{l|}{Our} & 25.5 \\ \bottomrule [1pt]
\multicolumn{2}{c|}{\textbf{Dry$\xrightarrow{}$ Rain}} & \multicolumn{2}{c}{\textbf{Day$\xrightarrow{}$ Night}} \\ \hline
\multicolumn{1}{l|}{DualCross~\cite{man2023dualcross}} & 29.6 &\multicolumn{1}{l|}{DualCross~\cite{man2023dualcross}} & 17.0 \\
\multicolumn{1}{l|}{PCT~\cite{pct}} & 27.2 & \multicolumn{1}{l|}{PCT~\cite{pct}} & 18.3 \\
\multicolumn{1}{l|}{Our} & 29.9 & \multicolumn{1}{l|}{Our} & 19.6 \\ \bottomrule [1pt]
\end{tabular}}
\end{center}
\vspace{-2em}
\end{table}
\begin{table}[t]
\fontsize{7}{7.8}\selectfont
\renewcommand{\arraystretch}{1.1}
\setlength\tabcolsep{9pt}
\caption{Vectorized HD mapping performance (mAP\%) on different UDA benchmarks.}
\vspace{-2.0em}
\label{tab:vecmapping}
\begin{center}
\resizebox{1.0\linewidth}{!}{
\begin{tabular}{lcccc}
\toprule [1.5pt]
\multicolumn{1}{l|}{\multirow{2}{*}{Method}}  & \multicolumn{3}{c|}{AP} & \multirow{2}{*}{mAP} \\ \cline{2-4}
\multicolumn{1}{l|}{}  & Bou. & Ped. & \multicolumn{1}{l|}{Div.} &  \\ \hline
\multicolumn{5}{c}{\textbf{Day $\xrightarrow{}$ Night}} \\ \hline
\multicolumn{1}{l|}{Source Only}  & 4.8 & 6.2 & \multicolumn{1}{l|}{10.8} & 7.3 \\
\multicolumn{1}{l|}{Domain ADV~\cite{man2023dualcross}}  & 5.4 & 5.3 & \multicolumn{1}{l|}{11.4} & 7.4 \\
\multicolumn{1}{l|}{Our}  & 5.4 & 6.1 & \multicolumn{1}{l|}{15.2} & \textbf{8.9} \\ \bottomrule [1pt]
\multicolumn{5}{c}{\textbf{Dry $\xrightarrow{}$ Rain}} \\ \hline
\multicolumn{1}{l|}{Source Only} & 36.9 & 38.7 & \multicolumn{1}{l|}{44.1} & 39.9 \\
\multicolumn{1}{l|}{Domain ADV~\cite{man2023dualcross}}  & 35.4 & 37.6 & \multicolumn{1}{l|}{41.8} & 38.3 \\
\multicolumn{1}{l|}{Our} & 37.1 & 39.0 & \multicolumn{1}{l|}{46.7} & \textbf{40.9} \\ \bottomrule [1.5pt]
\end{tabular}
}
\end{center}
\vspace{-2em}
\end{table}
\begin{table}[t]
\fontsize{9}{13.8}\selectfont
\renewcommand{\arraystretch}{1.1}
\setlength\tabcolsep{9pt}
\caption{Semantic HD mapping performance (mAP\%) on cross-dataset benchmarks.}
\vspace{-2em}
\label{tab:crossdata}
\begin{center}
\resizebox{1.0\linewidth}{!}{
\begin{tabular}{lllll}
\toprule [2pt]
\multicolumn{1}{l|}{\multirow{2}{*}{Method}} & \multicolumn{3}{c|}{IoU} & \multirow{2}{*}{mIoU} \\ \cline{2-4}
\multicolumn{1}{l|}{} & Boundary & Pedestrian & \multicolumn{1}{l|}{Divider} &  \\ \hline
\multicolumn{5}{c}{\textbf{nuScenes $\xrightarrow{}$ Argoverse}} \\ \hline
\multicolumn{1}{l|}{Source Only} & 8.5 & 0.6 & \multicolumn{1}{l|}{4.9} & 4.7 \\
\multicolumn{1}{l|}{DualCross~\cite{man2023dualcross}} & 12.0 & 0.0 & \multicolumn{1}{l|}{4.7} & 5.6 \\
\multicolumn{1}{l|}{Our} & 12.5 & 0.7 & \multicolumn{1}{l|}{10.8} & \textbf{8.0} \\ \bottomrule [2pt]
\multicolumn{5}{c}{\textbf{Argoverse $\xrightarrow{}$ nuScenes}} \\ \hline
\multicolumn{1}{l|}{Source Only} & 14.3 & 3.4 & \multicolumn{1}{l|}{10.5} & 9.4 \\
\multicolumn{1}{l|}{Our} & 15.4 & 4.1 & \multicolumn{1}{l|}{11.4} & \textbf{10.3} \\ \bottomrule [2pt]
\end{tabular}
}
\end{center}
\vspace {-1.0em}
\end{table}
\begin{table}[t]
\fontsize{7}{9.8}\selectfont
\renewcommand{\arraystretch}{1.1}
\setlength\tabcolsep{9pt}
\caption{The ablation result of the core module. It is evaluated in the setting of domain adaptation from $Botson \xrightarrow{} Singapore$.}
\label{tab:core}
\begin{center}
\resizebox{1.0\linewidth}{!}{
\begin{tabular}{cccccc}
\toprule [1.5pt]
MT &SGPS  & DACL & FXDA & CDFM & IoU \\ \hline
\checkmark   &  &  &  &  & 13.7 \\
\checkmark  & \checkmark &  &  &  & 15.8 \\
\checkmark  & \checkmark &  & \checkmark &  & 16.5 \\
\checkmark  & \checkmark & \checkmark & \checkmark &  & 16.7 \\
\checkmark  & \checkmark & \checkmark & \checkmark & \checkmark & 17.5\\ \bottomrule [1.5pt]
\end{tabular}
}
\end{center}
\vspace{-2em}
\end{table}
\textbf{Semantic Mapping}
We have selected two competitive domain adaptation methods for comparison: the adversarial learning from DualCross~\cite{man2023dualcross} and the perspective prior learning from PCT~\cite{pct}.
As shown in Table~\ref{tab:smapping}, the proposed method outperforms existing approaches in most domain gaps. 
Our method needs improvement in the low domain-gap rainy condition. This is because PCT uses a stronger baseline and higher resolution, which boosts accuracy but also requires a longer convergence time. The lack of publicly available code precluded a more detailed comparison.
Nonetheless, in the four cross-domain adaptation experiments, our method demonstrates improvements over the baseline by $+6.6\%$, $+5.7\%$, $+2.6\%$, and $+11.1\%$, respectively. 
As shown in Fig.~\ref{tab:car_iou}, even the auxiliary task of vehicle segmentation demonstrates the highest performance across all domain adaptation experiments,
confirming that our method not only performs well in mapping tasks but also exhibits corresponding capabilities in the BEV instance mask detection module.

\textbf{Vectorized HD Mapping}
For the vectorized mapping task, we similarly chose the adversarial learning method from work~\cite{man2023dualcross} as the baseline for comparison.
The domain adaptation results are shown in Table~\ref{tab:vecmapping}. 
As observed, our method exhibits superior performance, achieving accuracies of $8.9\%$ and $40.9\%$ in the cross-domain scenarios of $Day \xrightarrow{} Night$, and $Dry \xrightarrow{} Rain$, respectively.
It demonstrates that our method can be effectively and seamlessly integrated into domain adaptation learning for the vectorized mapping task.

\subsection{Cross-dataset Domain Adaptation Results}
The nuScenes and Argoverse datasets have vastly different sensor distribution deployments, which present a significant challenge for BEV adaptation learning. 
Therefore, we verify the effectiveness of our method in domain adaptation learning under the cross-dataset context.
Since the Argoverse dataset uses seven cameras while nuScenes utilizes six, we fix the number of cameras to six during domain adaptation training to ensure a fair comparison. 
Notably, for Argoverse, six perspectives are randomly selected for the perspective views. 
During evaluation, however, the original setup for each dataset is maintained.
As shown in Table~\ref{tab:crossdata}, our method significantly improves cross-dataset domain adaptation performance ($+2.3\%$) compared to the adversarial learning method~\cite{man2023dualcross} ($+0.9\%$). 
This indicates that the proposed model, coupled with hierarchical perspective priors, retains high-quality effectiveness in the cross-domain BEV mapping task.

\subsection{Ablation Results}
\textbf{Ablation of the Core Modules:}
HierDAMap builds upon the MT domain adaptation framework and innovatively proposes four core models. 
We will now sequentially explore the effectiveness of each module.
Table~\ref{tab:core} analyzes the effectiveness of different modules within HierDAMap. Initially, we implemented domain adaptation using the basic MT framework, which serves as our baseline. Subsequently, the supervision module with Semantic-Guided Pseudo Supervision (SGPS) provided a $+2.1\%$ improvement to the model. 
Further enhancing the model with the Feature Exchange Data Augmentation (FXDA) module with an additional $+0.7\%$ increase in accuracy.
Finally, the effectiveness of the core modules is validated with Dynamic-Aware Coherence Learning (DACL) and Cross-Domain Frustum Mixing (CDFM) modules. enabling the model to achieve the mapping accuracy of $17.5\%$.

\begin{figure}[tb]
      \centering
      \includegraphics[scale=0.55]{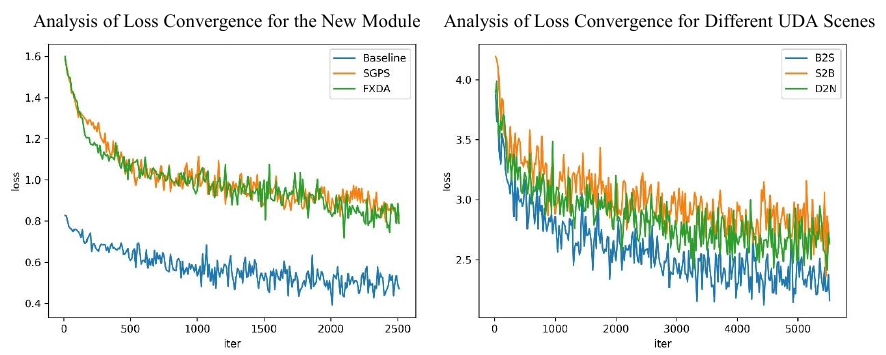}
      \vspace{-1.0em}
      \caption{The analysis of loss convergence for the new module and different UDA tasks.
      On the left, it illustrates the impact of the newly core loss components, SGPS and FXDA modules. On the right, across UDA tasks with different data types, the overall loss of HierDAMap converges properly.
      }
      \label{Fig.aba_loss}
      \vspace{-1.0em}
\end{figure}

Considering that core modules introduce additional loss terms to the training process, we further analyze the impact of core loss components on the overall convergence behaviour.
As shown in Fig.~\ref{Fig.aba_loss}, although the introduction of the SGPS module adds the $Loss_p$, the model still converges effectively. Moreover, the additional incorporation of the FXDA module further accelerates the convergence.

\begin{table}[t]
\fontsize{9}{13.8}\selectfont
\renewcommand{\arraystretch}{1.1}
\setlength\tabcolsep{9pt}
\caption{The ablation result of the auxiliary task in CDIG. It is evaluated in the setting of domain adaptation from $Botson \xrightarrow{} Singapore$. `A/M' means `Auxiliary task/Mixed Instance'.}
\vspace{-2.0em}
\label{tab:car}
\begin{center}
\resizebox{1.0\linewidth}{!}{
\begin{tabular}{ll|llllll|l}
\toprule [1.5pt]
A &M & Dri. & Ped. & Walk. & Stop. & Car. & Div. & mIoU    \\ \hline
 & & 50.0 & \textbf{11.1} & 14.8 & 5.2 & 5.1 & 14.1 & 16.7 \\
\checkmark & & 50.5 & 10.7 & \textbf{16.3} & 6.3 & 4.2 & 13.5 & 16.9  \\
\checkmark &\checkmark & \textbf{51.6} & 9.8 & 15.6 & \textbf{7.3} & \textbf{6.2} & \textbf{14.3} & \textbf{17.5}  \\ \bottomrule [1.5pt]
\end{tabular}}
\end{center}
\vspace{-2em}
\end{table}

\textbf{Ablation of Cross-Domain Instance Guidance:}
The relationship between dynamic vehicle instances and static maps is mutually reinforcing. 
This section further analyses how cross-domain dynamic instance mixing effectively enhances domain adaptation capabilities, as shown in Table~\ref{tab:car}. 
Firstly, we incorporated an instance detection auxiliary task. 
It shows a slight accuracy improvement, demonstrating the positive and beneficial role of dynamic instances in BEV mapping.
Then, by further incorporating the cross-domain instance guidance module, an overall improvement of $0.6\%$ in mapping accuracy was observed. 
Further analysis, although the accuracy for pedestrians and walkways slightly decreases, targets related to vehicle instances (such as drivable area, stop line, carpark area, and divider) show significant improvements in accuracy. 

\begin{figure}[t]
      \centering
      \includegraphics[scale=0.45]{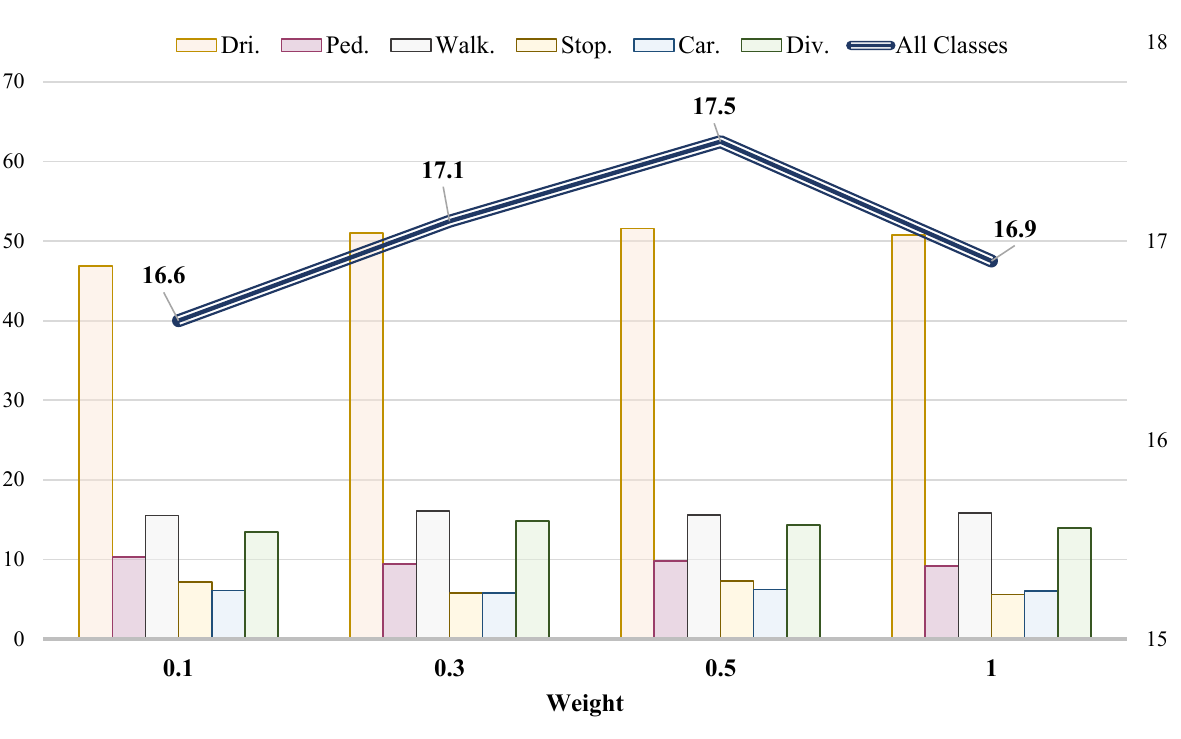}
      \vspace{-2em}
      \caption{The ablation result of different weights in SGPS. The model achieves optimal performance when the weight $\lambda_1$ is set to $0.5$.}
      \label{fig.pv_w}
      \vspace {-1.0em}
\end{figure}
\begin{table}[t]

\fontsize{9}{13.8}\selectfont
\caption{The ablation result of loss weight in FXDA. It is evaluated in the setting of domain adaptation from $Botson \xrightarrow{} Singapore$.}
\vspace{-1.0em}
\label{tab:weight}
\begin{center}
\begin{tabular*}{0.48\textwidth}{@{\extracolsep{\fill}}l|cccccc|c}
\toprule [2pt]
Weight & Dri. & Ped. & Walk. & Stop. & Car. & Div. &mIoU  \\ \hline
1$\times\beta$ &49.9 &9.5 &15.4 &5.7 &5.5 &14.8 &16.8 \\
2$\times\beta$ & 51.6 & 9.8 & 15.6 & 7.3 & 6.2 & 14.3 & 17.5 \\
5$\times\beta$ & 48.7 & 10.6 & 15.7 & 8.3 & 6.9 & 13.3 & 17.3 \\ \bottomrule [2pt]
\end{tabular*}
\end{center}

\caption{The ablation result of different View Transformers (VT). It is evaluated in the setting of domain adaptation from $Botson \xrightarrow{} Singapore$.}
\vspace{-1.5em}
\label{tab:vt}
\begin{center}
\begin{tabular*}{0.48\textwidth}{@{\extracolsep{\fill}}l|ccc|c}
\toprule [1.5pt]
VT & Boundary & Pedestrian & Divider &mIoU  \\ \hline
Fea. IPM~\cite{hdmapnet} & 17.7 & 1.5 & 16.7 & 11.9 \\
LSS & 20.1 & 1.5 & 18.1 & 13.2 \\ \bottomrule [1.5pt]
\end{tabular*}
\end{center}
\vspace {-2.0em}

\end{table}

\textbf{Ablation of Different Weights:}
In this section, we analyze the impact of two important weights.
The SGPS model supervises the image encoding layer through pseudo-labels, where the loss with different weights has a certain impact on the learning degree of image semantic features. 
Therefore, we first analyzed the weight influence of $\lambda_1$.
As shown in Fig.~\ref{fig.pv_w}, 
when the weight is $0.5$, the perspective images can maximally learn the semantic features required by BEV mapping.
Additionally, the proposed feature exchange data augmentation enhances efficiency by leveraging the strength of pseudo-label supervision loss from the target domain. 
To gain further insights, we analyze the impact of different loss weights on model learning, as shown in Table~\ref{tab:weight}. 
The results indicate that the module performs optimally when the weight multiplier is set to $2$. This finding aligns with the objective of using an auxiliary branch to enhance target domain learning efficiency, which requires a moderate weight to balance the auxiliary contribution.

\begin{figure*}[htb]
      \centering
      \includegraphics[scale=0.5]{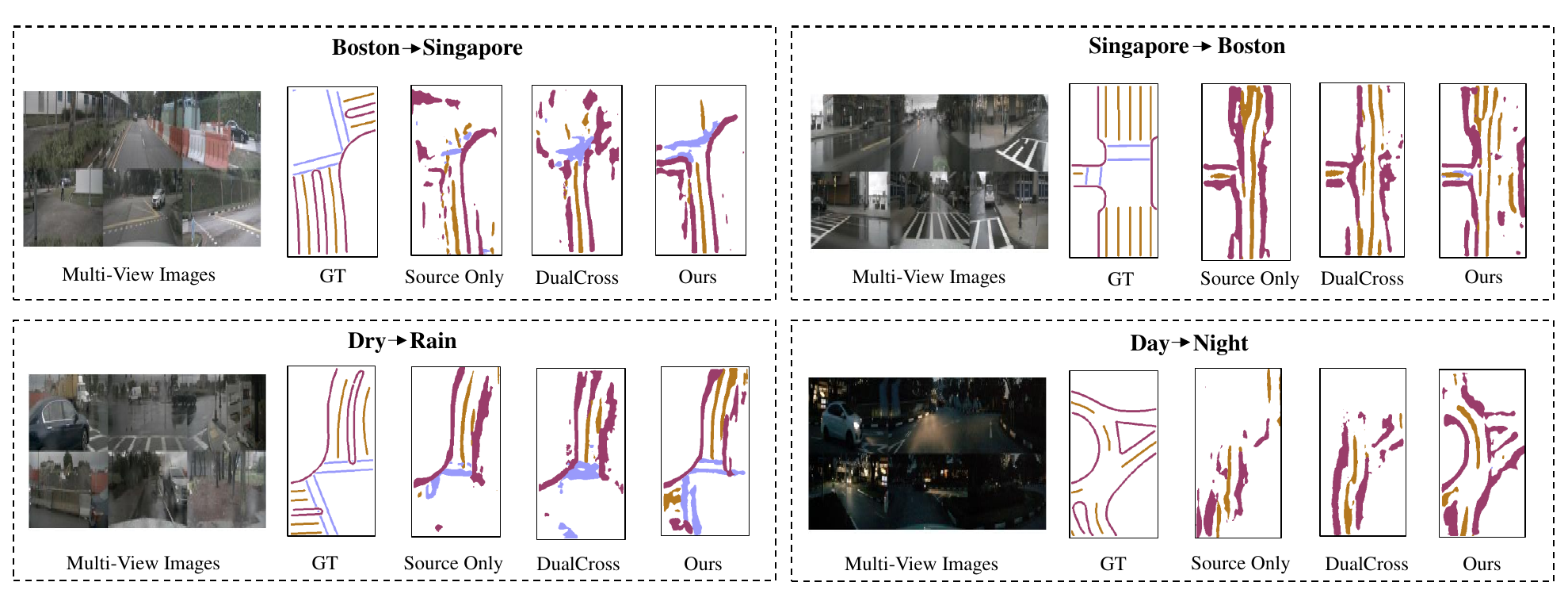}
      \vspace{-1.5em}
      \caption{Visualization results for semantic HD mapping. The proposed method is compared against the state-of-the-art method DualCross~\cite{man2023dualcross}. Classes of divider, pedestrian, and boundary are filled with red, blue, and yellow.}
      \label{fig.hdmap}
      \vspace {-1.5em}
\end{figure*}

\begin{table}[t]
\fontsize{6}{7}\selectfont
\renewcommand{\arraystretch}{1.1}
\setlength\tabcolsep{9pt}
\caption{The ablation result of different perspective prior models.}
\vspace{-2.0em}
\label{tab:pv_method}
\begin{center}
\resizebox{1.0\linewidth}{!}{
\begin{tabular}{l|cc}
 \toprule [1.0pt]
\multirow{2}{*}{Perspective Prior Models} & \multicolumn{2}{c}{mIoU} \\ \cline{2-3}
 & \multicolumn{1}{l|}{\textbf{Boston $\xrightarrow{}$ Singapore}} & \textbf{Day $\xrightarrow{}$ Night} \\ \hline
baseline & \multicolumn{1}{c|}{13.7} & --  \\
SAN~\cite{san} & \multicolumn{1}{c|}{17.5} & 19.9  \\
Mask2Former~\cite{mask2former} & \multicolumn{1}{c|}{15.4} &16.1  \\ \bottomrule [1.0pt]
\end{tabular}}
\end{center}
\vspace{-2em}
\end{table}
\textbf{Ablation of View Transformer:}
In addition to the LSS method employed in this paper, the Inverse Perspective Transformation (IPM) method is also widely used in BEV mapping tasks, owing to its strong generalization capability.
Therefore, we analyzed the effectiveness of different view transformer methods, as shown in Table~\ref{tab:vt}.
In this section, we chose the IPM method~\cite{hdmapnet}, which operates at the feature level, as the compared method. 
Note that, aside from the view transformer module, the other modules remain consistent.
The results show that the LSS with depth supervision demonstrates stronger applicability under the same sensor configuration.

\textbf{Ablation of Different Perspective Prior Models:}
Different perspective prior models exhibit varying levels of accuracy. To fully validate the effectiveness of the proposed method, this section conducts evaluations under different models.
As shown in Table~\ref{tab:pv_method}, the perspective priors of SAN and Maskformer models can be effectively applied to the hierarchical domain adaptation framework. Among them, SAN demonstrates stronger guidance capabilities.
Meanwhile, this section also validates the effectiveness of different domain adaptation methods under various perspective prior settings. The PCT work uses perspective priors as pseudo-labels to achieve supervision. Since the source code for PCT is not publicly available, this section instead incorporates a semantic pseudo-label supervision module into the UniMatch~\cite{unimatch} and MT frameworks, respectively, to serve as baseline comparison methods.
As shown in Table~\ref{tab:imgsize}, the proposed hierarchical domain adaptation framework achieves the highest accuracy under both perspective prior models, with improvements of $2.3\%$ mIoU and $0.5\%$ mIoU, respectively.

\begin{table}[t]
\fontsize{6}{7.5}\selectfont
\renewcommand{\arraystretch}{1.1}
\setlength\tabcolsep{9pt}
\caption{The ablation result of image size and perspective prior models. $*$ means the results are from the referenced work. $^{\dagger}$  represents the replicated results in this paper.}
\vspace{-2.0em}
\label{tab:imgsize}
\begin{center}
\resizebox{1.0\linewidth}{!}{
\begin{tabular}{lccc}
\toprule [1.5pt]
\multicolumn{1}{l|}{Method}    & \multicolumn{1}{l|}{Image Size} & \multicolumn{1}{l|}{Perspective Prior Models} & mIoU \\ \midrule [1pt]
\multicolumn{4}{c}{\textbf{Day $\xrightarrow{}$ Night}}                    \\ \hline
\multicolumn{1}{l|}{DualCross~\cite{man2023dualcross}} & \multicolumn{1}{l|}{128${\times}$352}    & \multicolumn{1}{c|}{--}                 & 15.5 \\
\multicolumn{1}{l|}{PCT*~\cite{pct}}      & \multicolumn{1}{l|}{128${\times}$352}    & \multicolumn{1}{c|}{--}                & 16.9 \\
\multicolumn{1}{l|}{PCT-UniMatch$^{\dagger}$~\cite{unimatch}}      & \multicolumn{1}{l|}{128${\times}$352}    & \multicolumn{1}{c|}{Mask2Former}                 & 15.6 \\
\multicolumn{1}{l|}{HierDAMap} & \multicolumn{1}{l|}{128${\times}$352}    & \multicolumn{1}{c|}{Mask2Former}                 & 16.1 \\ 
\multicolumn{1}{l|}{PCT-UniMatch$^{\dagger}$~\cite{unimatch}}      & \multicolumn{1}{l|}{128${\times}$352}    & \multicolumn{1}{c|}{SAN}                 &17.0  \\
\multicolumn{1}{l|}{PCT-MT$^{\dagger}$}      & \multicolumn{1}{l|}{128${\times}$352}    & \multicolumn{1}{c|}{SAN}                & 17.6 \\
\multicolumn{1}{l|}{HierDAMap} & \multicolumn{1}{l|}{128${\times}$352}    & \multicolumn{1}{c|}{SAN}                 & 19.9 \\ 

\bottomrule[1.5pt]
\end{tabular}}
\end{center}
\vspace{-2em}
\end{table}

\textbf{Accuracy Analysis under the Same Image Size:}
This subsection presents a comparative analysis of the accuracy of various UDA methods under identical resolution conditions.
As shown in Table~\ref{tab:imgsize}, both the PCT and the proposed approach, which incorporate prior knowledge, demonstrate excellent performance. Notably, the proposed hierarchical prior framework achieves even stronger results.

\begin{table}[tb]
\fontsize{4.5}{6.5}\selectfont
\renewcommand{\arraystretch}{1.1}
\caption{Comparison of GFLOPs and the number of parameters (\#Params) between the baseline and our method.}
\begin{center}
\label{table:realtime}
\resizebox{1.0\linewidth}{!}{
\begin{tabular}{l|c|cc}
\toprule [1pt]
Method    &Perspective Prior & GFLOPs  & \#Params \\ \midrule[0.5pt]
Baseline  & & 50.97G & 9.23M \\
Our-Eval  & & 50.97G & 9.23M \\
Our-Train &\checkmark & 24.23G & 9.46M \\ \bottomrule[1pt]
\end{tabular}}
\end{center}
\end{table}
\begin{table}[t]
\fontsize{9}{13.8}\selectfont
\renewcommand{\arraystretch}{1.1}
\setlength\tabcolsep{9pt}
\caption{Vector HD mapping performance (mAP\%) with different target sensors.}
\vspace{-2.0em}
\label{tab:tarsensors}
\begin{center}
\resizebox{1.0\linewidth}{!}{
\begin{tabular}{l|llc}
\toprule [2pt]
Method & Source Sensors & Target Sensors & mAP \\ \hline
Source Only & Camera + LiDAR & -- & 7.3 \\ \hline
\multirow{2}{*}{Domain ADV} & Camera + LiDAR  & Camera & 7.4 \\
 & Camera + LiDAR  & Camera+LiDAR & 14.9 \\ \hline
\multirow{2}{*}{Ours} & Camera + LiDAR  & Camera & 8.9 \\
 & Camera + LiDAR  & Camera+LiDAR & 15.5 \\ \bottomrule [2pt]
\end{tabular}
}
\end{center}
\vspace {-2.0em}
\end{table}
\textbf{Analysis about Computational Cost and Real-time Efficiency:}
Real-time performance is a critical requirement for the practical deployment of BEV models. 
To fully leverage hierarchical perspective priors for enhanced model generalization, this paper introduces several novel modules. 
Therefore, this subsection provides a comprehensive analysis of computational cost and real-time capability.
As shown in Table~\ref{table:realtime}, while the combination of modules such as the perspective view segmentation branch and auxiliary prediction heads increases the parameter count during training, these branches can be discarded during inference.  Consequently, the computational cost of our method during validation is identical to that of the baseline.
In other words, this approach incurs no additional overhead to the real-time speed during inference. 

\textbf{Discussion of Different Target Sensors:}
The depth values from LiDAR sensors can provide a significant positive influence on BEV model learning, particularly enhancing the accuracy of estimating spatial relationships. This leads us to consider whether different sensor configurations in the target domain might impact cross-domain learning.
We tested the vectorized mapping adaptation in the target domain with LiDAR depth supervision, as shown in Table~\ref{tab:tarsensors}.
Interestingly, although our method improves the adaptation capability under pure vision conditions, the addition of depth supervision in the target domain results in a substantial boost in model accuracy, which further enhances our model's domain adaptation accuracy by $+7.6$ mAP.
This underscores the importance of learning geometric spatial relationships in the view transformer module for BEV domain adaptation research.

\begin{table}[tb]
\fontsize{4}{4.5}\selectfont
\renewcommand{\arraystretch}{1.1}
\caption{Experimental results of domain mixing schemes in different directions.}
\vspace{-1.5em}
\begin{center}
\label{tab:path}
\resizebox{1.0\linewidth}{!}{
\begin{tabular}{lcc|c}
\toprule [0.8pt]
CDFM    &source $\xrightarrow{}$ target &target $\xrightarrow{}$ source  & mIoU \\ \midrule[0.1pt]
\multicolumn{4}{c}{\textbf{Day $\xrightarrow{}$ Night}}                    \\ \hline
\checkmark  &\checkmark &  & 19.9 \\
\checkmark  &\checkmark &\checkmark &19.3  \\ \bottomrule[0.8pt]
\multicolumn{4}{c}{\textbf{Boston $\xrightarrow{}$ Singapore}}                    \\ \hline
\checkmark  &\checkmark &  & 17.5 \\
\checkmark  &\checkmark &\checkmark &16.5  \\ \bottomrule[0.8pt]
\end{tabular}}
\end{center}
\vspace{-1.5em}
\end{table}
\begin{figure*}[htb]
      \centering
      \includegraphics[scale=0.5]{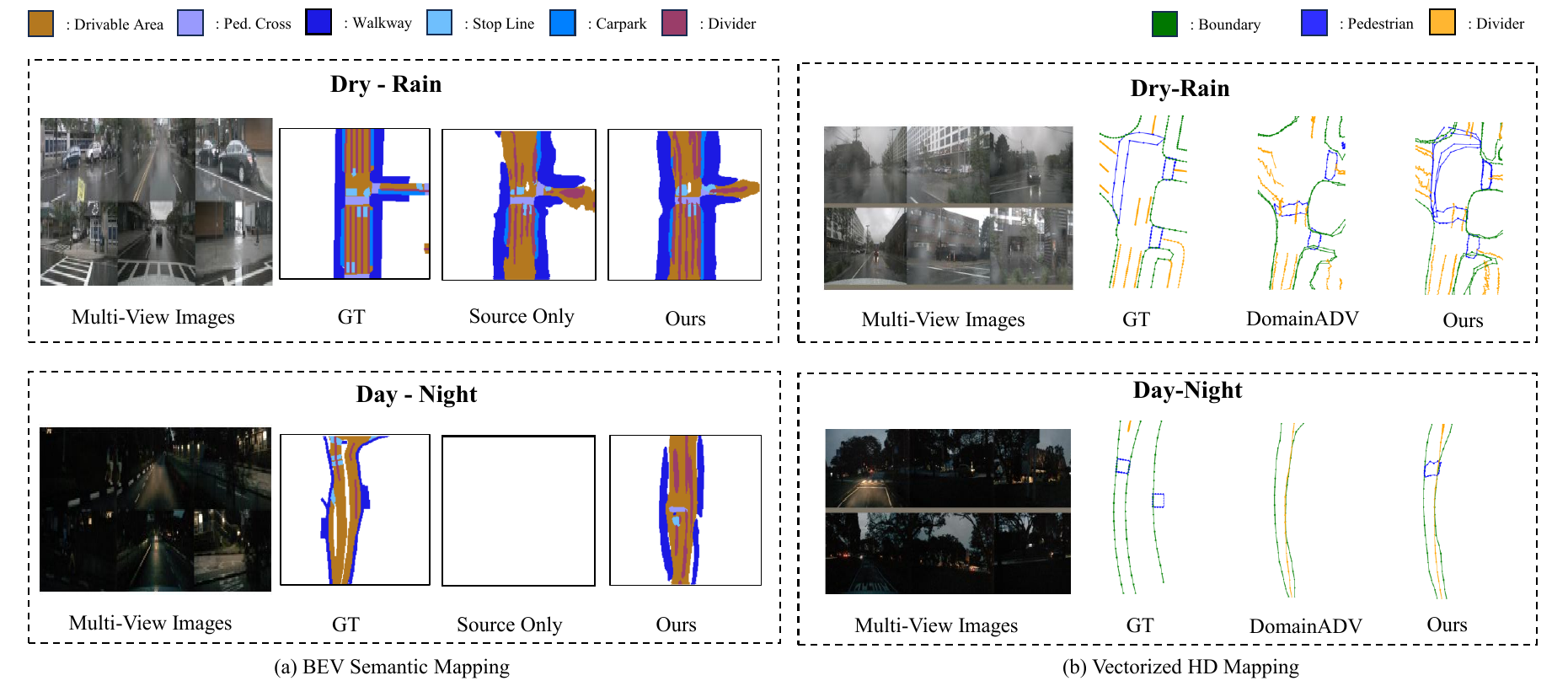}
      \vspace{-0.45cm}
      \caption{Visualization results for semantic mapping and vectorized HD mapping. It presents a visual comparison of our method against the source-only method and the adversarial learning approach~\cite{man2023dualcross}. The source domain model for BEV semantic mapping is initially unsuitable for night conditions, but after domain adaptation learning, it gains a better understanding of the environment.}
      \label{fig.svmap}
      \vspace {-1.0em}
\end{figure*}
\textbf{Discussion of Domain Mixing Paths:}
Our domain mixing strategy adopts a source-to-target direction, aiming to reduce the distribution discrepancy while preserving reliable source-domain supervision. Considering recent bidirectional domain mixing strategies in image-based domain adaptation, we further investigate their applicability to BEV perception.
Specifically, we introduce an additional target-to-source mixing branch to construct a bidirectional mixing scheme. The results are reported in Table~\ref{tab:path}. The bidirectional strategy does not improve cross-domain performance, indicating that it is not directly suitable for BEV adaptation. Unlike 2D image tasks, target-domain BEV representations involve accumulated geometric and semantic uncertainties during view transformer and pseudo label generation. The reverse mixing therefore introduces unreliable spatial structures and conflicting supervision signals, degrading the stability of BEV feature learning.
Designing effective multi-directional domain mixing strategies that consider BEV geometric consistency remains an interesting direction for future research.

\subsection{Visualization Analyses}
Fig.~\ref{fig.hdmap} illustrates the cross-domain visualization performance of DualCross~\cite{man2023dualcross} and our method across different domain distributions on the nuScenes dataset. 
It is evident that compared to the adversarial learning strategy of DualCross, the model proposed in this study is more capable of delineating the map, particularly demonstrating more effective mapping of pedestrians in rainy conditions. 
The visualization results of BEV semantic mapping and vectorized mapping are shown in Fig.~\ref{fig.svmap}. 
On one hand, it demonstrates the effectiveness of our model; on the other hand, it shows that our method outperforms purely adversarial learning approaches.
Especially in the BEV segmentation mapping of Fig.~\ref{fig.svmap}, compared to the baseline, which shows no effectiveness, our method can provide superior semantic mapping capabilities in the setting of domain adaptation from $Day \xrightarrow{} Night$.
Simultaneously, our method can more accurately depict map instances in vectorized mapping tasks.

\section{Conclusion}
In this paper, we propose HierDAMap, a universal domain adaptation framework based on hierarchical perspective priors for various BEV map construction tasks. 
Driven by visual foundational models, this paper thoroughly explores the guiding learning capabilities of perspective priors at three levels: global semantics, sparse class, and individual instances.
At the global level, supervision is directly applied to image encoding through perspective pseudo labels. 
The sparse level employs dynamic-aware dynamic labels generated from perspective pseudo-labels and predicted depth distributions to enforce consistency in the process of the view transformer. 
The instance level utilizes perspective instance masks to implement a domain mixing strategy, simultaneously generating BEV instance labels based on the corresponding view frustum in BEV space.
Our proposed method is rigorously evaluated across six cross-domain benchmarks within two datasets and three distinct tasks, consistently achieving state-of-the-art performance. Visualization analyses further corroborated that our approach exhibits superior adaptability for diverse BEV mapping tasks.

\textbf{Limitation:} 
In domain mixing tasks, ensuring the plausibility of instances remains a research challenge. In the proposed CDFM module, spatial discrepancies between different domains may lead to implausible placement of mixed instances. 
For example, when distant instances from the source domain are blended into target domain images, they may appear to float on the surface of nearby obstacles rather than being occluded behind them as required by spatial geometry, resulting in physically inconsistent compositions.
This represents a limitation that we aim to address in future work.

\textbf{Future Research Directions:}
We aim to explore the capabilities of prior knowledge, like temporal and depth information, to further enhance the domain generalizability of BEV mapping.
Furthermore, by leveraging temporal trajectory information of instances, the domain mixing strategy can analyze the spatial relationships between instances and the target domain, thereby enhancing the reliability of domain mixing.

\bibliographystyle{IEEEtran}
\bibliography{refer.bib}

\end{document}